\documentclass{article}
\usepackage{ijcai20}
\usepackage{general}

\usepackage{times}
\usepackage{bbm}

\usepackage{soul}
\usepackage{url}
\usepackage[hidelinks]{hyperref}
\usepackage[utf8]{inputenc}
\usepackage[small]{caption}
\usepackage{graphicx}
\usepackage{psfrag}
\usepackage{pstool}
\usepackage{amsmath}
\usepackage{booktabs}
\newcommand{\Omegan}{\Omega^{(n)}}
\newcommand{\Omegam}{\Omega^{(m)}}
\newcommand{\Omegak}{\Omega^{(k)}}

\newcommand{\Qn}{Q^{(n)}}

\newcommand{\Qm}{Q^{(m)}}
\newcommand{\Qk}{Q^{(k)}}
\newcommand{\Pn}{P^{(n)}}
\newcommand{\Pk}{P^{(k)}}

\newcommand{\indicator}[1]{{\mathbbm 1}[#1] }

\newcommand{\tauind}{\tau_{\rm ind}}

\newcommand{\pointmass}[1]{ 1_{#1} }

\def\defterm#1{\emph{#1}}

\title{A Complete Characterization of Projectivity for Statistical Relational Models
  \footnote{This Paper consists of an unaltered version of the paper published at IJCAI 2020 to which
  an appendix has been added. The bibliography is extended with a few additional references used in the appendix.  }}

 \author{
 Manfred Jaeger$^1$\footnote{Contact Author}\And
 Oliver Schulte$^2$
 \affiliations
 $^1$Computer Science Department, Aalborg University, Aalborg, Denmark\\
 $^2$School of Computing Science, Simon Fraser University, Burnaby, Canada
 \emails
 jaeger@cs.aau.dk, oschulte@cs.sfu.ca
 }
\begin{document}

\maketitle

\begin{abstract}
A generative probabilistic model for relational data consists of a family of probability distributions for relational structures over domains of different sizes. In most existing statistical relational learning (SRL) frameworks, these models are not projective in the sense that the marginal of the distribution for size-$n$ structures on induced substructures of size $k < n$ is equal to the given distribution for size-$k$ structures. Projectivity is very beneficial in that it directly enables lifted inference and statistically consistent learning from sub-sampled relational structures. In earlier work some simple fragments of SRL languages have been identified that represent projective models. However, no complete characterization of, and representation framework for projective models has been given. In this paper we fill this gap: exploiting representation theorems for infinite exchangeable arrays we introduce a class of directed graphical latent variable models that precisely correspond to the class of projective relational models. As a by-product we also obtain a characterization for when a given distribution over size-$k$ structures is the statistical frequency distribution of size-$k$ substructures in much larger size-$n$ structures. These results shed new light onto the old open problem of how to apply Halpern et al.'s ``random worlds approach'' for probabilistic inference to general relational signatures.
\end{abstract}

\section{Introduction}

Many types of generative models have been proposed for relational data in several fields, including machine learning and statistics.
For i.i.d. data, a parametrized model defines a distribution over samples of a fixed size $n$, for every $n$. The analogue for generative relational models is a distribution $\Qn$ over complex multi-relational graphs (``worlds'' in logical terminology) of a fixed size $n$, for every $n$. Research in statistical theory and discrete mathematics on the one hand, and AI and machine learning on the other hand has focussed on somewhat different aspects of relational models: the former is mostly concerned with internal model properties such as exchangeability, projectivity and behavior in the limit, whereas the latter is focussed on learning and inference tasks for one size $n$ at a time.  

It is well known that in many popular statistical relational learning (SRL) frameworks the dependence of $\Qn$ on $n$  exhibits sometimes counter-intuitive and hard to control behavior.  Most types of SRL models are not projective 
in the sense that the distribution $\Qn$ for $n$ nodes is the marginal distribution derived from the $Q^{n+1}$ distribution~\cite{Shalizi2013,Jaeger2018}. For exponential random graph and Markov logic network (MLN) models it has also been observed that the $\Qn$ tend to become degenerate as $n$ increases in the sense that the probability becomes concentrated on a few ``extreme'' structures~\cite{rinaldo2009,chatterjee2013estimating,poole2014population}. Some authors have proposed to better control the behavior of MLNs by adjusting the model parameters as a function of $n$~\cite{jain2010adaptive}; however, no strong theoretical guarantees have yet been derived for such approaches. 

In this paper we focus on projectivity as a very powerful condition to control the behavior of $\Qn$. In projective models, inferences about a fixed set of individuals are not sensitive to population size. This implies that inference trivially becomes \emph{domain-lifted}~\cite {broeck2011completeness}, convergence of query probabilities becomes trivial, and certain statistical guarantees for learning from sub-sampled relational structures can be obtained~\cite{Jaeger2018}. These benefits come at a certain cost in terms of expressivity: projective models are necessarily ``dense'' in the sense that, e.g., the expected number of edges in a projective random graph model is quadratic in $n$. In spite of these limitations, there exist  projective model types such as the stochastic block model and the infinite relational model~\cite{xu2006learning,kemp2006learning} that have been proven very useful in practice. It thus seems very relevant to fully exploit the capabilities of projective models by developing maximally expressive projective representation, learning and inference frameworks. In this paper we take an important step in this direction by deriving a complete characterization of projective models as a certain class of directed latent variable models.   

While the characterization we obtain is completely general, we approach our problem from the perspective that knowledge about the distributions $\Qn$ is given in the form of
statistical frequencies of substructures of a small size $k$. For example, $k$ could be the maximal number of variables in an MLN formula, in which case the substructure frequencies are a sufficient statistics for learning the MLN parameters. In a somewhat different setting, $k$ can be the number of variables used in a Halpern/Bacchus-style statistical probability formula forming a statistical knowledge base~\cite{Halpern90,Bacchus90}. In all cases the question arises of how to generalize this knowledge to  infer probabilities for specific instances (``beliefs"), either by statistical model estimation (as in most current SRL frameworks), or by inferring plausible beliefs based on invariance or maximum entropy principles, as in the random worlds approach of Bacchus et al.~\shortcite{BaGroHalKol92}, and more recently in~\cite{kern2010novel} and~\cite{kuzelka2018relational}. A fundamental question that then arises is whether the given substructure frequencies can actually be the marginal distribution of $\Qn$ for large $n$. Results about the random worlds method need to be conditioned on the assumption that the statistical knowledge is ``eventually consistent''~\cite[Chapter 11]{halpern2017reasoning}. Similar assumptions are made in~\cite{kuzelka2018relational}. As a by-product of our characterization of projective models we obtain that the same characterization also describes the distributions that can be induced as marginals of arbitrary $\Qn$.

\section{Related Work} We discuss work on generative graph models related to exchangeability and projectivity, the two key properties in our study.
\paragraph{Exchangeability.}  Exchangeability requires that a generative model should assign the same probability to graphs that differ only in node labellings. This is true for the large class of template-based relational models, because typical model discovery methods do not introduce templates that  reference  individual nodes~\cite{Kimmig2014}. For example, they may only construct first-order logic formulas with no constant symbols. This includes most structure learning algorithms for Markov Logic Networks (e.g., \cite{Schulte2012}).\footnote{An exception is the Boostr system \cite{Khot2013}, which constructs first-order MLN formulas with constants.} Similarly, the sufficient statistics of exponential random graph models (e.g., the number of triangles in a graph) are typically defined without special reference to any particular node.  Niepert and Van den Broeck~\shortcite{niepert2014tractability} have exploited the weaker notion of \emph{partial exchangeability} to obtain tractable inference for certain SRL models.

\paragraph{Projectivity.} The importance of projectivity for graph modelling has been discussed previously~\cite{Shalizi2013,Jaeger2018}. Chatterjee and Diaconis~\shortcite{chatterjee2013estimating} discuss how estimation and inference in exponential random graph models depends on the sample size.  Shalizi and Rinaldo~\shortcite{Shalizi2013} give necessary and sufficient projectivity conditions for an exponential random graph model; they show that these are satisfied only in rare conditions. Jaeger and Schulte~\shortcite{Jaeger2018}  discuss a number of common SRL models, including MLNs and Relational Bayesian Networks, and show that they are projective only under restrictive conditions. Projective models  used in practice factor a graph into independent components given a set of latent variables. Popular examples include the stochastic block model and generalizations~\cite{hoff2002latent}, the infinite relational model~\cite{Orbanz2014}, and recent graph neural network models such as the graph variational auto-encoder~\cite{Kipf2016}. Our work shows that a latent conditional independence representation is not only sufficient for projectivity, but also necessary. We prove this result for a very large class of structured data,  essentially general finite multi-dimensional arrays (tensors) with no restrictions on their dimensionality.
Our results heavily depend on the theory of infinite exchangeable multi-dimensional arrays~\cite{hoover1979relations,aldous1981representations,kallenberg2006probabilistic,Orbanz2014}. 
The question of realizability of a given frequency distribution as a relational marginal has also been raised by Kuzelka et al.\shortcite{kuzelka2018relational}, who then focus on approximate realizability, rather than characterizations of exact realizability.

\section{Background}

\subsection{Basic Definitions}

We use the following basic notation. The set of integers $\{1,\ldots,n\}$ is denoted $[n]$.
For any $d\geq 1$, we write
$[n]_{\neq}^d$ for the set of $d$-tuples containing $d$ distinct elements from $[n]$.  
The subset of $[n]_{\neq}^d$ containing tuples in which the elements appear in their
natural order is denoted
$\langle n \rangle^d$ (so that $\langle n \rangle^d$
corresponds to a standardized representation for the set of all $d$-element subsets of $[n]$).
Extending this notation to the infinite case, we can also write $[\Nset]_{\neq}^d$ and
$\langle \Nset \rangle^d$.

\paragraph{Relations and Possible Worlds.}
A relational \emph{signature} $S$\ contains relations
of varying arities. We 
refer to the maximal arity of relations contained in $S$ as the \emph{arity of} $S$, denoted
$\emph{arity}(S)$. 
A \emph{possible world} $\omega$\ (for $S$) specifies  
1) a finite domain $D=\{d_1,\ldots,d_n\}$, 2) 
for each $m$-ary relation from $S$ an $m$-dimensional binary adjacency matrix.
We  refer to $n$ as the \emph{size} of 
$\omega$, and also call $\omega$ an $n$-world. For most purposes, we can assume that $D=[n]$, or
at least $D\subset \Nset$. However, even if we make this assumption for convenience of presentation,
we do not generally assume that the
integer label of a randomly observed domain element can also be observed. 
We  denote by $\Omegan$ the set of 
all possible worlds for a given signature $S$ with domain $[n]$. The relevant signature
is usually implicit from the context, and not made explicit in the notation. 
Finally, $\Omega:=\cup_n \Omega^{(n)}$. 

\begin{figure}
  \centering
  \includegraphics{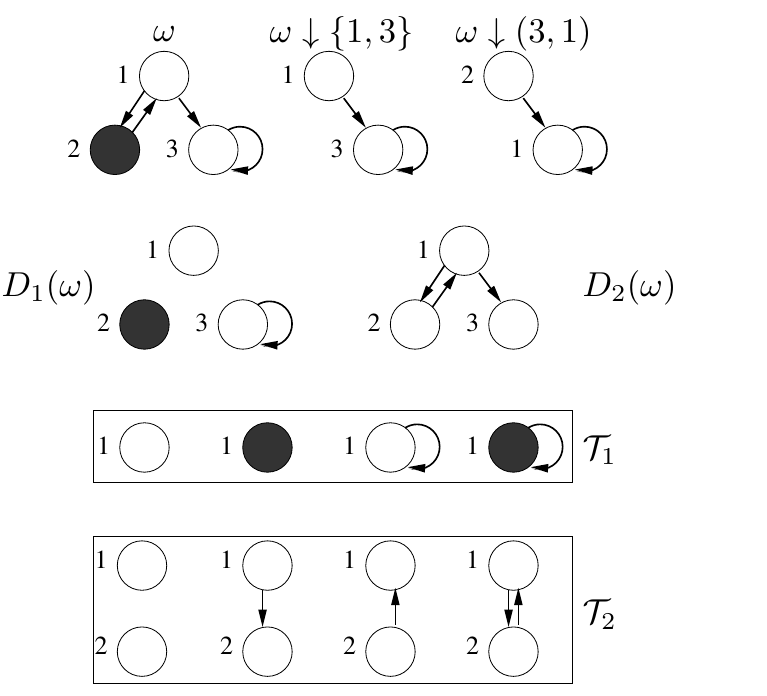}
    \caption{Top left: world $\omega$ with one unary relation (black/white) and one
      binary (edge) relation; top middle/right: sub-worlds induced by $I=\{1,3\}$ and
      $\boldi=(3,1)$;
      second row: unary and binary data parts; bottom: spaces ${\cal T}_1,{\cal T}_2$ for
    the given signature.}
  \label{fig:worlddefs}
\end{figure}

\paragraph{Relational Substructures.}
We also require notation to refer to different types of substructures of a possible $n$-world $\omega$. For
a subset $I\subset [n]$ of size $|I|=m<n$ we denote with $\omega\downarrow I$ the $m$-world induced by
$I$, i.e., the possible world with domain $I$, and the relations of $\omega$ restricted to arguments from $I$.
For a tuple $\boldi\in [n]^m_{\neq}$ we denote with $\omega\downarrow \boldi$  the world
over the domain $[m]$ obtained by relabeling the domain elements in the sub-world induced by the
set $\boldi$ as $i_h\mapsto h$  (cf. Figure~\ref{fig:worlddefs}, top row). 
A little less conventional is the following concept, that will become important for our main theorem: for
$m=1,\ldots,  \emph{arity}(S)$ we define $D_m(\omega)$ as the \emph{arity-$m$ data of $\omega$}. Informally speaking,
$D_m(\omega)$ collects all the information from all adjacency arrays of $\omega$ that refers to
exactly $m$ distinct elements. For example (cf. Figure~\ref{fig:worlddefs}), $D_1(\omega)$ contains the
data (adjacency arrays) of all unary relations of $S$, but also the information contained on the diagonal
of a two-dimensional adjacency array for a binary (edge) relation, i.e., the information about self-loops
of that relation. A possible world can then also be described by the tuple $(D_m)_{m=1,\ldots,\emph{arity}(S)}$.
Furthermore, $D_m(\omega)$ can be decomposed into the factors $D_m(\omega\downarrow \boldi)$, where
$\boldi$ ranges over $\langle n\rangle^m$. We denote with ${\cal T}_m$ the space of possible
values of $D_m(\omega\downarrow \boldi)$ ($|\boldi|=m$). A possible world $\omega\in\Omegan$ then also is given
by an assignement of a value in ${\cal T}_m$ for all $\boldi\in\langle n\rangle^m$ ($m=1,\ldots, \emph{arity}(S)$).

\section{Worldlet Frequency Distributions}
\label{sec:worldletfreqs}


Many graph analysis methods examine frequent characteristic subgraphs to provide information about a larger graph. We can think of a subgraph as a template that can be instantiated multiple times in a large graph. For example, in a social network we can count the number of friendship triangles among women. Depending on the framework, such templates go by different names (e.g., graphlets, motifs, frequency subgraphs) and are represented using different syntax (e.g., SQL queries, first-order logic, semantic relationships). We observe that subgraph templates can be represented in a general syntax-independent way as the collection of fully specified graphs $\Omegak$ of a fixed size $k$, where we think of $k$ as a small number (typically in the range $k=2,\ldots,5$). When seen as a subgraph pattern, we  refer to a world $\omega \in \Omegak$ as a \defterm{worldlet}.
We assume that for every worldlet, the frequency of its occurrence in a larger world is available, through learning or expert elicitation (cf. \cite{Bacchus90}). As a notational convention, we use $k$ and $n$ to denote domain sizes of (small) worldlets and large ``real'' worlds, respectively. This convention  only is intended to support intuitions,
and does not have any strict mathematical implications.

\paragraph{Statistical Frequency Distributions.}

The intuitive idea of observing random worlds by sampling subsets of larger domains
can be formalized in slightly different ways, e.g. by  assuming sampling with or without replacement, or
by interpreting the observation as a unique world, or only an isomorphism
class~\cite{diaconis2007graph,kuzelka2018relational}. In many aspects alternative sampling models
become essentially equivalent as $n\rightarrow\infty$~\cite{diaconis2007graph}.
We here adopt a sampling model in which an ordered sample is drawn without replacement.
Thus, a sample from a world $\omega\in\Omegan$ is given by one of the $n!/(n-k)!$ tuples
$\boldi\in[n]^k_{\neq}$, and the observed worldlet then is $\omega\downarrow\boldi$.
Note that this sampling method does not rely on observing the original labels of elements drawn from
$\omega$ to obtain the labeling of elements in the sampled worldlet, and therefore also makes sense when
the elements of $\omega$ can not be assumed to have (observable) integer labels.
The frequency distribution obtained through this sampling method is denoted ${P}^{(k)}(\cdot|\omega)$.

  \begin{example}
    Let $S=\{e\}$ consist of a single binary relation. Let $\omega\in\Omegan$ be a ``star'' with center 1, i.e.,
    $e$ consists of the edges $\{1\rightarrow l: l=2,\ldots,n\}$. The probability that a random draw of 2 elements
    contains the node 1 then is $2/n$, with equal probability that 1 is the first or second drawn element.
    The three worldlets
    $1\bullet\!\! {\white\rightarrow} \!\! \bullet 2$,
    $1\bullet\!\! \rightarrow\!\! \bullet 2$ and $1\bullet\!\! \leftarrow \!\! \bullet 2$
    then have probabilities $1-2/n, 1/n,  1/n$ (in this order) 
     under ${P}^{(k)}(\cdot|\omega)$.
  \end{example}


  Every world $\omega$ defines a  frequency distributions ${P}^{(k)}(\cdot|\omega)$.
  If first a random $\omega$ is selected, we obtain a two-step sampling procedure that was first described
  in a more general context by Fenstad~\shortcite{Fenstad67}. 

\paragraph{Fenstad Sampling.}
Given a possible world distribution $\Qn$, we define the \defterm{expected statistical frequency}
distribution $\Pk\circ\Qn$ for $k$-worlds $\omega'$ as follows:
\begin{equation}
  \label{eq:twostepsampling}
  (\Pk\circ\Qn) (\omega') := \sum_{\omega\in\Omegan} \Qn(\omega)\Pk(\omega'\mid\omega).
\end{equation}
We denote with $\Delta^{(k)}_{n}$ the set of distributions on $\Omegak$ that have a representation
of the form (\ref{eq:twostepsampling}) for some $\Qn$. If $k<l<n$, then 
$\Pk\circ(P^{(l)}\circ\Qn)=\Pk\circ\Qn$, and thus  $\Delta^{(k)}_{n}\subseteq \Delta^{(k)}_{l}$.

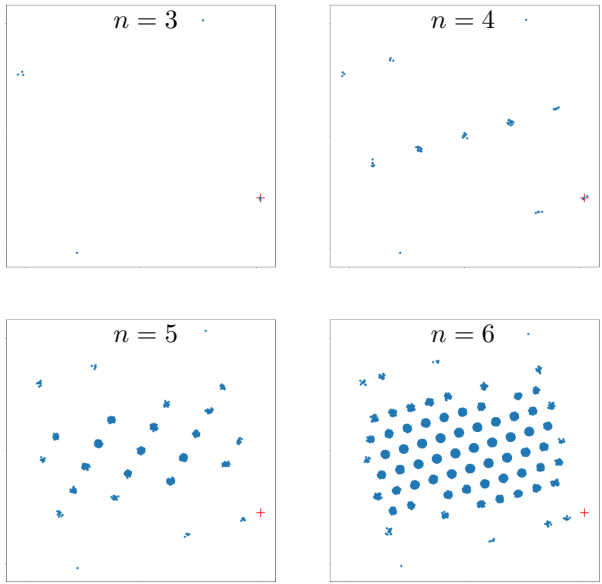
\begin{figure}
  \centering
  \includegraphics[scale=0.2]{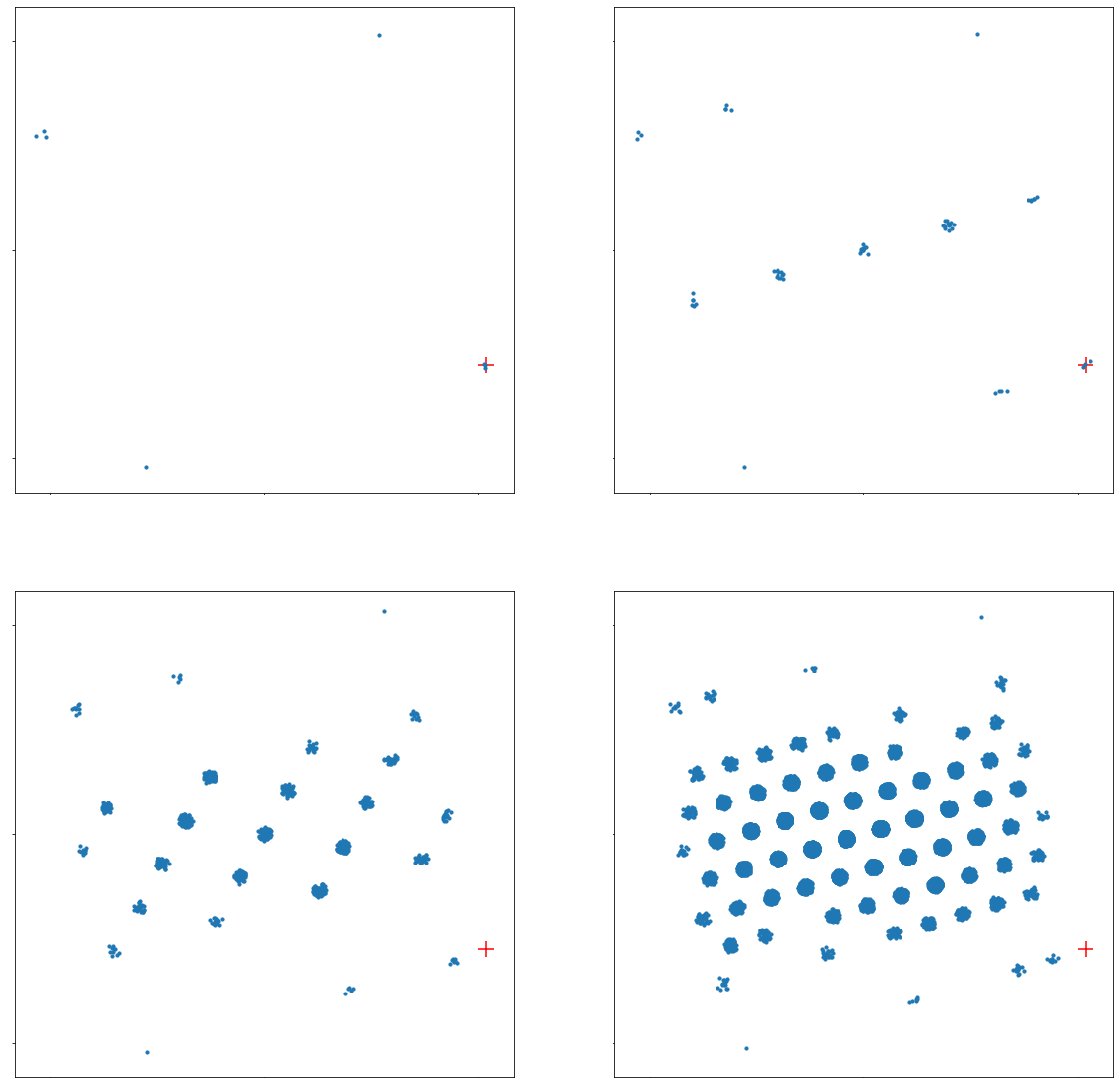}
  \put(-185,212){$n=3$}
  \put(-65,212){$n=4$}
  \put(-185,93){$n=5$}
  \put(-65,93){$n=6$}
  \caption{Illustration of $\Delta^{(k)}_{n}$ for $k=3$ and $n=3,4,5,6$. Cf. examples~\ref{ex:Deltakn} and
    \ref{ex:running1}}
  \label{fig:scatterfig}
\end{figure}

\begin{example}
  \label{ex:Deltakn}
  In this example and some of the following, we take $S$ to contain a single undirected edge relation $e$. In order
  to comply with our general definitions, which are based on directed relations, we consider an undirected edge
  $i\bullet - \bullet j$ to be a shorthand for the conjunction $i\bullet \rightarrow \bullet j$ and
  $i\bullet \leftarrow \bullet j$, and we assume that all worlds with uni-directional edges
  ($i\bullet \rightarrow \bullet j$ but not
  $i\bullet \leftarrow \bullet j$) or self-loops ( $i\bullet \rightarrow \bullet i$) have probability zero.
  Disregarding these probability zero worlds, $\Omega^{(3)}$ then contains 8 possible worlds belonging to
  4 different isomorphism classes. The top row of Table~\ref{tab:wdistributions} depicts these isomorphism
  classes, together with the count of worlds in each class.

  Figure~\ref{fig:scatterfig} illustrates for $n=3,4,5,6$ the worldlet frequency distributions $\Pk(\cdot|\omega)$ defined
  by the worlds $\omega\in\Omegan$. Each (blue) dot is the distribution defined by one world
  after projecting its 8-dimensional probability vector into 2-dimensional space. Some jitter is applied to
  exhibit the multiplicities of $n$-worlds defining the same distribution on worldlets of size 3.
  The sets $\Delta^{(k)}_{n}$ are the convex hulls of these points. The distribution marked by the (red) + in
  Table~\ref{tab:wdistributions} and Figure~\ref{fig:scatterfig} belongs to
  $\Delta^{(k)}_{n}$ for $n=3,4$, but not for $n=5,6$.
\end{example}

\section{Relational Models and Distribution Families}
\label{sec:properties}


As our goal is to examine properties of relational models that are independent of a particular model syntax, we use a  family of distributions as a semantic view of a parametrized model. The two key properties of families in our study are exchangeability and projectivity.

\subsection{Distribution Families: Exchangeability, and Projectivity}

\begin{definition}

A \emph{family of distributions} $\{\Qn: n \in\Nset\}$ specifies, for each finite domain size $n$, a distribution $\Qn$ on the possible world set $\Omegan$. 
\end{definition}

\begin{definition}
A probability distribution $\Qn$ on $\Omegan$ is \emph{exchangeable}, if
  $\Qn(\omega)=\Qn(\omega')$ whenever $\omega$ and $\omega'$ are isomorphic.
  A family is exchangeable, if every member of the family is exchangeable. 
\end{definition}

Intuitively a distribution family is projective if its members are mutually consistent in the sense that the world distribution over a smaller domain size is the marginal distribution over a larger one. For a precise definition, we 
follow our notation for relational substructures, and for each $n$-world $\omega$, write $\omega \downarrow [m]$ for the size-$m$ subworld that results from restricting $\omega$ to the first $m$ elements. 
A distribution $\Qn$ over $n$-worlds then induces a {\em marginal} probability for an $m$-world $\omega'$ as follows: 

\begin{equation*}
\Qn \downarrow[m] (\omega') = \sum_{\omega \in \Omegan: \omega \downarrow [m] = \omega'}\!\!\!\!\!\! \Qn(\omega) 
\end{equation*}

Projectivity is the central concept for our investigation:
\begin{definition}
  \label{def:projective}
  An exchangeable family $(\Qn)_{n\in\Nset}$ is \emph{projective}, if for all
  $m<n$: $\Qn\downarrow [m] =\Qm$.
\end{definition}
Note that  in contrast to more
general notions of projectivity found in the theory of stochastic processes,
we here define projectivity only for exchangeable families. Exchangeability implies that
the marginal distribution $\Qn\downarrow I$ is the same for all subsets $I$ of size $m$, and therefore
we only need to consider the  marginal $\Qn\downarrow [m]$ as a prototype.

\begin{example}
  Statistical frequency distributions  $\Pk(\cdot\mid\omega)$ always are
  exchangeable. As a special case, if $\omega\in\Omegan$, then  $\Pn(\cdot\mid\omega)$ samples a random
  permutation of $\omega$, i.e., is the uniform distribution on the isomorphism class of $\omega$.
  It follows that distributions
  defined by Fenstad sampling (\ref{eq:twostepsampling}) also are exchangeable, for any
  $\Qn$. 
\end{example}


We approach the question of how to characterize and represent projective families through the more
specific question of whether a given distribution $\Qk$ can be embedded in a projective family. The
following definition provides the necessary terminology.

\begin{definition}
  \label{def:extendable}
  Let  $\Qk$ be an exchangeable distribution on $\Omegak$. $\Qk$ is called
  \begin{itemize}
  \item \emph{$n$-extendable}, if $\Qk\in\Delta^{(k)}_{n}$; any $\Qn$ that
    induces $\Qk$ via (\ref{eq:twostepsampling}) is called an \emph{extension} of  $\Qk$.
  \item \emph{extendable}, if it is $n$-extendable for all $n>k$;
  \item \emph{projective extendable} if there exists a projective family $(\Qn)_n$ of extensions
    of $\Qk$.
  \end{itemize}
\end{definition}




\begin{table}
  \centering
  \begin{tabular}{cccc|p{10mm}}
    \includegraphics[scale=0.2]{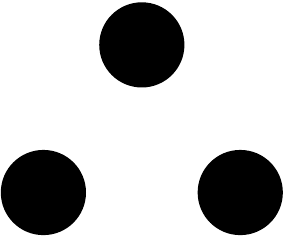} ($\times 1$)&
    \includegraphics[scale=0.2]{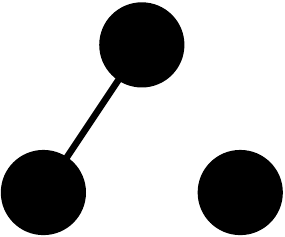} ($\times 3$)&
    \includegraphics[scale=0.2]{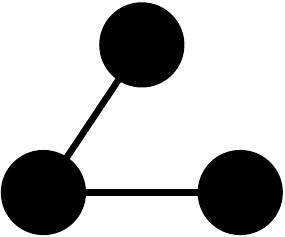} ($\times 3$)&
    \includegraphics[scale=0.2]{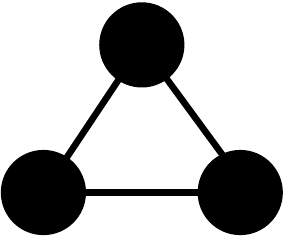} ($\times 1$)& Name \\
    \hline
    1 & 0 & 0 & 0 & $\pointmass{E_3}$ \\
    0 & 0 & 0 & 1 & $\pointmass{K_3}$ \\
    0 & 1/3 & 0 & 0 & + \\
    1/4 & 0 & 1/4 & 0 & bipart\\ 
  \end{tabular}
  \caption{Some example worldlet distributions}
  \label{tab:wdistributions}
\end{table}

\begin{example}
\label{ex:running1}
  The rows in Table~\ref{tab:wdistributions} specify several exchangeable distributions on $\Omega^{(3)}$ (in the
  undirected graph setting, as described in Example~\ref{ex:Deltakn}). The numbers in the table specify the probabilities
  of each world in a given isomorphism class, not the total probability of the isomorphism class.
  The first two are the point masses on the
  empty graph (denoted $E_3$) and complete graph ($K_3$), respectively. If $\pointmass{E_n}$ denotes the point mass
  on the empty graph of size $n$, then $( \pointmass{E_n})_n$ is a projective family. Similarly for the
  family $(\pointmass{K_n})_n$, and the family of mixtures  $(0.5\cdot\pointmass{E_n}+0.5\cdot\pointmass{K_n})_n$.

  The row labeled + is the distribution marked by the (red) + in the plots of Figure~\ref{fig:scatterfig}.
  If $\omega\in\Omega^{(4)}$ is the graph that contains the two edges $1\bullet\!\!-\!\!\bullet 2$ and 
  $3\bullet\!\!-\!\!\bullet 4$, then this
  distribution is equal to $P^{(3)}(\cdot|\omega)$. Thus, it is 4-extendable, which is also visible in the top right
  panel of Figure~\ref{fig:scatterfig} showing that '+' coincides with sampling distributions induced by
  4-worlds. 
  However, '+' is not $n$-extendable for any  $n\geq 5$. 
  This is visible in Figure~\ref{fig:scatterfig} as for $n=5,6$  '+'  lies outside the convex hull
  of the worldlet frequency distributions. Proposition~\ref{prop:modularity} below will provide a simple tool
  for proving the non-extendability of '+'. 

  The last row in the table describes the distribution that in the limit for $n\rightarrow \infty$ is the
  worldlet frequency distribution defined by complete, balanced bipartite graphs, i.e., graphs whose edge set
  is equal to $\{i\bullet\!\!-\!\!\bullet j: 1\leq i\leq \lfloor n/2 \rfloor; \lfloor n/2 \rfloor +1\leq j \leq n  \}$. It will follow
  from our main theorem that this distribution is projective extendable.
\end{example}

\subsection{Domain Sampling Distributions}

Extendable distributions $\Qk$ in the sense of Definition~\ref{def:extendable} are mixtures of worldlet frequency
distributions. An important special case is when $\Qk$  is a pure worldlet frequency distribution
$\Pk(\cdot|\omega)$ defined by a single world $\omega$. In that case, however, one cannot expect that $\Qk$
can be represented in this form with suitable $\omega$ for all $n$, because the sets
$\{\Pk(\cdot|\omega): \omega\in\Omegan \}$ for different $n$ are concentrated on different
grids of rational numbers, and therefore are largely disjoint (cf. Figure~\ref{fig:scatterfig}).
Following the approach already taken by Bacchus et al. to give semantics to statistical probability terms
in the random worlds approach~\cite{BaGroHalKol92,halpern2017reasoning}
we therefore only require that $\Qk$ is approximately equal to some $ \Pk(\cdot|\omega)$, with an increasing
accuracy in the approximation as the size of $\omega$ increases.




\begin{definition}
\label{def:dsrealizable}
  Let   {$\Qk$} be a probability distribution on 
$\Omegak$. We say that {$\Qk$}  is  a \emph{domain sampling distribution} 
if the following holds: 
for every $\epsilon >0$ there exists $n\in\Nset$, such that for every $n'\geq n$: there exists
a possible $n'$-world $\omega$, so that for all $\omega'\in\Omegak$:
\begin{equation}
  \label{eq:realizable}
  |P^{(k)}(\omega'\mid \omega)-\Qk(\omega')|<\epsilon.
\end{equation}
\end{definition}

Thus, the property of being a domain sampling distribution strengthens the property
of extendability in that in the representation (\ref{eq:twostepsampling})
only point masses
$\Qn=\pointmass{\omega}$ are allowed, but weakens it in that (\ref{eq:realizable}) only
requires approximate equality.

\begin{example}
  For the worldlet distributions of Table~\ref{tab:wdistributions} we have
  $\pointmass{E_3}=P^{(3)}(\cdot| E_n)$ for all $n\geq 3$,
  so that $\pointmass{E_3}$ is a domain sampling distribution
  (with zero approximation error). Similarly for $\pointmass{K_3}$. The mixture
  $0.5\cdot \pointmass{E_3}+0.5\cdot \pointmass{K_3}$ is projective extendable, but not a domain
  sampling distribution. The distribution '$+$' is not a domain sampling distribution. This is
  indicated by Figure~\ref{fig:scatterfig}, because already for $n=6$ the distribution is separated
  by a  distance $\epsilon>0$ from the set $\Delta^{(3)}_6$. Because of the nested structure of the
  $\Delta^{(3)}_n$ there then also cannot be better approximations for larger $n>6$. The last
  'bipart' distribution in Table~\ref{tab:wdistributions} again is a domain sampling distribution
  with a non-zero approximation error that only vanishes as $n\rightarrow\infty$.
\end{example}

\section{A Representation Theorem}
\label{sec:reptheo}


We now proceed to derive our main result, which is a comprehensive characterization of
families $(\Qn)_n$ and worldlet marginals $\Qk$ with the structural properties
described in Section~\ref{sec:properties}. 
We introduce a representation for projective families that is based on the analysis and
representation theorems for infinite exchangeable arrays developed by
Aldous~\shortcite{aldous1981representations} and  Hoover~\shortcite{hoover1979relations}. The definitive treatment is given
by Kallenberg~\shortcite{kallenberg2006probabilistic}. We therefore call the following an AHK model.

\begin{definition}
  \label{def:ahkmodel}
  Let $S$ be a signature with maximal $\emph{arity}(S)=a\geq 1$. An AHK model for $S$ is given by
  \begin{itemize}
  \item A family of i.i.d. random variables $\{U_{\boldi}|  \boldi\in \langle \Nset \rangle^m, m=0,\ldots,a\}$,
    where each $U_{\boldi}$ is uniformly distributed on $[0,1]$.
  \item A family of random variables  $\{D_{\boldi}|  \boldi\in \langle \Nset \rangle^m, m=1,\ldots,a\}$. 
    For $\boldi\in \langle \Nset \rangle^m$ the variable $D_{\boldi}$ takes values in ${\cal T}_m$.
  \item For each $m=1,\ldots,a$ a measurable function
    \begin{equation}
      \label{eq:ffunction}
      f^m: [0,1]^{2^m}\rightarrow {\cal T}_m
    \end{equation}
    so that
    \begin{itemize}
    \item for $\boldi=(i_1,\ldots,i_m)\in\langle \Nset \rangle^m$ the value of $D_{\boldi}$ is defined as
      $f^m(\boldU_{\boldi})$, where 
  \begin{multline}
    \label{eq:Tfromf}
    \boldU_{\boldi}=(U_{\emptyset},U_{i_1},\ldots,U_{i_m},U_{( i_1,i_2)},\ldots,\\
    U_{( i_{m-1},i_m)},\ldots
    \ldots, U_{( i_1,\ldots,i_m) }),
  \end{multline}
 is the vector containing all $U_{\boldi'}$-variables with $\boldi'\subseteq\boldi$  in
  lexicographic order.
\item $f^m$ is permutation equivariant, in the sense that for any permutation $\pi$ of $[m]$
  \begin{displaymath}
    f^m(\pi\boldU_{\boldi})=\pi f^m(\boldU)
  \end{displaymath}
  where $\pi\boldU_{\boldi}$ is the permutation of $\boldU_{\boldi}$ that in the place of
  $U_{\boldi'}$ contains $U_{\pi\boldi'}$ with $\pi\boldi'$ the ordered tuple of the elements
  $\{\pi(i): i\in\boldi'\}$.
    \end{itemize}
  \end{itemize}

  An AHK model that does not contain the $U_{\emptyset}$ variable is called an AHK$^-$ model.
\end{definition}

\begin{figure}[tb]
  \centering
 \includegraphics{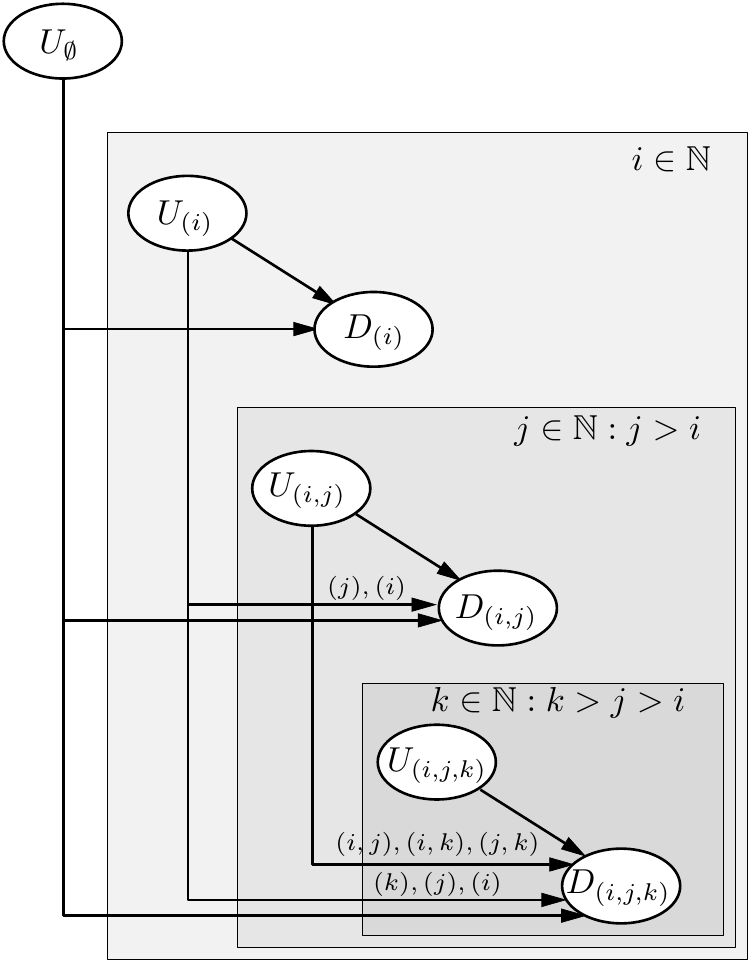}
  \caption{Plate representation of AHK model with $a=3$}
  \label{fig:ahkplates}
\end{figure}

Figure~\ref{fig:ahkplates} gives an illustration of the structure of an AHK model in plate notation.
  An AHK model is fully determined by the functions
  $\boldf:=(f^m)_{m=1\ldots,a}$, and we therefore write $\boldf$ to refer to an AHK model.
  By a slight abuse of notation, we also use $\boldf$ to denote the distribution
  defined by the model  on the possible worlds over the infinite domain $\Nset$, and
  write  $\boldf\downarrow [n]$ for the marginal on the induced sub-world  over
  the domain $[n]$.

  The following example gives a simple illustration of how the permutation equivariance condition for the 
functions $f^m$ ensures exchangeability.

    \begin{example}
      We encode a version of the Erd\H{o}s-R\'enyi random graph model in which any pair of nodes is 
connected with probability 1/2 by an edge, and that edge is given a random direction. Thus, the target distribution on worldlets 
of size 2 is $P(1\bullet \leftarrow \bullet 2)=P(1\bullet \rightarrow \bullet 2)=0.25$,
$P(1\bullet \hspace{3mm} \bullet 2)=0.5$.
The state space ${\cal T}_1$ contains the
  two states ``self-loop'' and ``no self-loop''. Since self-loops have probability zero, we simply
  let $f^1$ be the constant function that returns ``no self-loop'' regardless of the input $U$-variables.
  The state space ${\cal T}_2$ contains the four states 
  $1\bullet\!\! {\white\rightarrow} \!\! \bullet 2$,
  $1\bullet\!\! \rightarrow\!\! \bullet 2$, $1\bullet\!\! \leftarrow \!\! \bullet 2$,
  and $1\bullet\!\! \leftrightarrow \!\! \bullet 2$, of which only the first three have non-zero probability.
Let
\begin{multline*}
  f^2(x_0,x_1,x_2,x_3):= \\
  \left\{
    \begin{array}{ll}
      1\bullet \rightarrow \bullet 2 & \mbox{if}\ x_1 < x_2 \ \mbox{and}\ x_3 <0.5 \\
      1\bullet \leftarrow \bullet 2 & \mbox{if}\ x_2 < x_1 \ \mbox{and}\ x_3 <0.5 \\
      1\bullet \hspace{3mm} \bullet 2 & \mbox{otherwise.}
    \end{array}
    \right.
  \end{multline*}
  For clarity we here use a notation that makes it clear that the functions $f^m$ are defined on
  arrays of length $2^m$, and their definition distinguishes arguments by their position in the input
  array, not by their semantic nature as a variable $U_{\boldi'}$. 
  For $\pi: 1\mapsto 2, 2\mapsto 1$ we then have $\pi(1\bullet \rightarrow \bullet 2 )= 1\bullet \leftarrow \bullet 2$,
  and $f^2(\pi\boldU_{(1,2)})=f^2(U_{\emptyset},U_2,U_1,U_{(1,2)})= \pi f^2(\boldU_{(1,2)})$. Together with the fact that
  the tuples $\boldU_{(1,2)}$ and $\pi\boldU_{(1,2)}$ have identical distribution, this implies that
  the two values $ 1\bullet \rightarrow \bullet 2$, $ 1\bullet \leftarrow \bullet 2$ of $D_{(1,2)}$ have the same
  probability.
    \end{example}

    Generalizing from this example, and also noting that the plate representation of the AHK models
    directly implies that  marginals $\boldf\downarrow [n]$ simply are given  by
instantiating the plate model only for $\boldi\subset [n]$, we can note the following proposition.

\begin{proposition}
  \label{prop:ahkprojective}
    Let $\boldf$ be an AHK model.
    The marginals $\boldf\downarrow [n]$ are exchangeable, and the family
    $(\boldf\downarrow [n])_n$ is projective.
  \end{proposition}

 
For a given worldlet distribution $\Qk$ with $k\geq \emph{arity}(S)$ we say that $\Qk$
\emph{has an AHK representation}, if there exists an $\boldf$ with $\boldf\downarrow [k]=\Qk$.

We can now formulate our main result.

\begin{theorem}
  \label{theo:extendable2}
Let $\Qk$ be an exchangeable distribution on $\Omega^{(k)}$ with $k\geq \emph{arity}(S)$. For the statements
  
\begin{description}
  \item[(A)] $\Qk$ is a domain sampling distribution.
  \item[(B)] $\Qk$ has a AHK$^-$ representation
  \item[(C)] $\Qk$ is a finite mixture of domain sampling distributions  
  \item[(D)] $\Qk$ is  extendable
  \item[(E)] $\Qk$ is  projective extendable
  \item[(F)] $\Qk$ has a AHK representation
  \end{description}

  the following implications hold:

  \begin{displaymath}
    \mbox{\bf (A)}\Leftrightarrow  \mbox{\bf (B)}
    \Rightarrow \mbox{\bf (C)}
    \Leftrightarrow \mbox{\bf (D)}
    \Leftrightarrow \mbox{\bf (E)}
    \Leftrightarrow \mbox{\bf (F)}
  \end{displaymath}

\end{theorem}

The full proof of the theorem is given in the extended online version of this paper (\url{http://arxiv.org/abs/2004.10984}). 
\section{Discussion}

\begin{figure}
  \centering
  \includegraphics{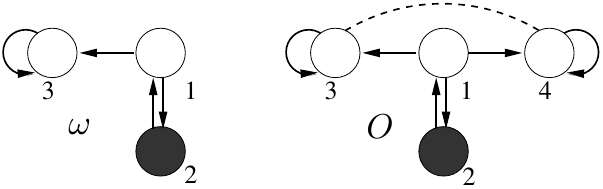}
  \caption{Modularity of AHK models: if $\omega$ on the left has nonzero probability, then also
  the set of worlds $O$ on the right.}
  \label{fig:modular}
\end{figure}

In this section we consider some of the trade-offs between limitations in expressivity of
projective models on the one hand, and
gain in algorithmic and statistical tractability on the other hand.
Limitations in expressivity can be considered in terms of what distributions $\Qn$, for a fixed $n$, can
be represented, and in terms of  the limitations for the family $\{\Qn|n\in\Nset\}$ as a whole.
Considering a single distribution $\Qn$, we can observe a \emph{modularity} property as described
by the following proposition, and illustrated in Figure~\ref{fig:modular}

\begin{proposition}
  \label{prop:modularity}
  Let $\boldf$ be an AHK model, $\omega\in\Omegan$ with $\boldf\downarrow[n](\omega)>0$.
  Let $O\subset\Omega^{(n+1)}$ be the set of $n+1$ worlds $\omega'$ for which
  $\omega'\downarrow [n]=\omega'\downarrow \{1,\ldots,n-1,n+1 \} =\omega$.
  Then $\boldf\downarrow[n](O)>0$. Moreover, if $\boldf$ is an AHK$^-$ model, then
  $\omega'\downarrow [n]=\omega$ and $\omega'\downarrow \{1,\ldots,n-1,n+1 \} =\omega$ are
  independent events given $\omega'\downarrow [n-1]$.
\end{proposition}

Figure~\ref{fig:modular} illustrates the proposition with $n=3$: if the world $\omega$ on the left
has nonzero probability, then also the set of 4-worlds $O$ on the right has nonzero probability.
$O$ is the set of 4-worlds for which the substructures induced by
$\{1,2,3\}$ and $\{1,2,4\}$ are both isomorphic to $\omega$.  The dashed
arc connecting nodes 3 and 4 on the right indicates that the value of $D_{(3,4)}$ determining the
relations between nodes 3 and 4 can vary for
different elements of  $O$.

As an application of Proposition~\ref{prop:modularity} we can see that  the '+' distribution  of Table~\ref{tab:wdistributions}
does not have an AHK representation, and therefore cannot be extendable (cf. Example~\ref{ex:running1}):
letting $n=2$ and $\omega= 1\bullet\!\!-\!\!\bullet2$, we obtain from the proposition that also 3-worlds with
two edges $1\bullet\!\!-\!\!\bullet2$ and $2\bullet\!\!-\!\!\bullet3$ must have nonzero probability, which is not
the case for '+'.


We now turn to structural limitations of the whole family $\{\Qn|n\in\Nset\}$ implied by an AHK representation.
As already 
mentioned in the introduction, projective families generate structures that are ``dense'' in the limit.
More precisely, if $\omega\in\Omegak$ is a worldlet with $\boldf\downarrow [k] (\omega)>0$, then
the expected number of $k$-tuples  in worlds of size $n$ which induce sub-worlds isomorphic to $\omega$ grows
linearly in $n^k$. Specifically, if graph edges have a nonzero probability at all, then the expected
number of edges grows linearly in $n^2$. It must be emphasized, though, that this only imposes limits on
modeling the asymptotic behavior of evolving graphs. For any fixed domain size, an AHK model can fit 
any observed degree distribution:


\begin{example}
  Let $n^*\in\Nset$, and  let $f(d)$ ($d=0,1,\ldots n^*$) denote an out-degree distribution  for
  directed graphs on $[n^*]$. 
  For arbitrary $n$ we can normalize out-degrees in graphs of size $n$ via  $d\mapsto d/n$.
  Let $F(\delta)$ ($\delta\in[0,1]$) be the cumulative distribution function obtained from $f()$ 
  for the normalized degrees $d\mapsto d/n^*$. We now define
    \begin{multline*}
    f^2(U_i,U_j,U_{(i,j)}):= \\
    \left\{
      \begin{array}{ll}
        i\bullet\!\! \rightarrow \!\! \bullet j & \mbox{if}\ U_i\geq F(U_{(i,j)})\ \mbox{and}\ U_j < F(U_{(i,j)})  \\
        i\bullet\!\! \leftarrow \!\! \bullet j & \mbox{if}\ U_j\geq F(U_{(i,j)})\ \mbox{and}\ U_i < F(U_{(i,j)})   \\
        i\bullet\!\! \leftrightarrow \!\! \bullet j & \mbox{if}\ U_i\geq F(U_{(i,j)})\ \mbox{and}\ U_j \geq F(U_{(i,j)})   \\
        i\bullet\!\! {\white\leftrightarrow} \!\! \bullet j & \mbox{otherwise}
      \end{array}
      \right.
    \end{multline*}
    Let $\delta_i$ denote the normalized out-degree of node $i$.
    Then for all $u\in[0,1]$ we obtain the expected normalized out-degree:
    \begin{equation}
      \label{eq:expoutdeg}
      E[\delta_i|U_i=u]=F^{-1}(u).
    \end{equation}
    $U_i$ being uniformly distributed, the right-hand side of (\ref{eq:expoutdeg}) is distributed with cdf
    $F()$, and so the expected normalized degree distribution follows $F()$. In the special case $n=n^*$ then
    the expected absolute degree distribution is the original $f()$.
\end{example}

On the positive side, we obtain significant computational and robustness advantages from the use of projective
models: inference is \emph{lifted} in the strongest possible sense that the complexity of computing a query
probability for a query involving $k$ named entities is independent of the size of the domain in which
the entities are embedded. For learning, projectivity is a necessary condition for consistent estimation from
substructures randomly sampled from domains of unknown size. However, further conditions beyond projectivity
are required to formulate and derive precise consistency guarantees~\cite{Jaeger2018}.
Statistical consistency and robustness results can therefore not be directly given for AHK models in general
without first identifying a suitable effectively representable and parameterizable class of functions
from which the $f^m$ can be constructed. Identifying rich and tractable such classes, and evaluating their
learning capabilities empirically and theoretically is future work. 

When evaluating the trade-offs of AHK models for a particular application, it must always be born in mind
that the strenghts of generative, projctive models only come to bear when one needs to deal with diverse
types of queries (so that a discriminative model for a fixed prediction task would be inadequate), and
when one has to deal with data from domains of different and/or uncertain sizes. We note that this is basically
the opposite side of the task spectrum from where many current popular node classification and link
prediction problems are situated, in which both learning and inference is conducted for a fixed task on
a single given graph, e.g., \cite{wu2020comprehensive}.

\section{Conclusion}

In this paper we have laid theoretical foundations for the study and application of 
rich classes of projective families.
Bringing together research strands in statistical graph theory and statistical relational learning
we have derived an explicit characterization of projective families in the form of a directed
graphical (plate) model. We have shown that closely linked to projectivity is the (approximate)
realizability as a statistical frequency distributions of worldlet samples drawn from large domain.
These results give us  a characterization of the form of statistical knowledge to which
the random worlds approach of Bacchus et al.~\shortcite{BaGroHalKol92} can be applied.

Interestingly, the structure of
AHK models has much in common with the ``independent choice logic'' family of SRL
frameworks~\cite{Sato95,Poole97,KimDemDeRSanRoc11} that also generate random relational structures as
deterministic functions of a set of a-priori independent random variables. However, the continuous
nature of the $U_{\boldi}$ variables in the AHK model, and the potential need of functions $f^m$ not readily
expressible in existing SRL languages pose significant challenges for the direct application of existing
SRL techniques.

On the theoretical side, many interesting questions remain regarding statistical principles of model
selection, and unbiasedness and consistency of estimation: for a given worldlet distribution $\Qk$ there
will often be multiple AHK models that precisely fit $\Qk$ and therefore are indistinguishable based on
likelihood scores. What invariance, parsimony, or plain parameter regularization principles are
then most useful for model selection?

\section*{Acknowledgments} Oliver Schulte's contribution was supported by a Discovery Grant from the Natural Sciences and Engineering Research Council of Canada. 

\vspace{5mm}

\appendix

{\parindent 0mm
{\bf \LARGE Appendix}
}
\section{Details on ${\cal T}_m$, $D_{\boldi}$}

We fully formalize the definitions of ${\cal T}_m$, $D_m$ and $D_{\boldi}$.

Let $r\in S$ be a $k$-ary relation. The interpretation of $r$ in a possible $n$-world $\omega$ is given
by a $k$-dimensional 0,1-valued adjacency matrix $A_r(\omega)$, which we view as a mapping
$[n]^k\rightarrow \{0,1\}$. Let $m\leq k$, and denote by $[n]^k_m$
the subset of $[n]^k$ that contains all $k$-tuples with exactly $m$ distinct elements. Let
$A_r(\omega)|m$ denote the restriction of $A_r(\omega)$ to tuples from  $[n]^k_m$. Then
\begin{displaymath}
  D_m(\omega) :=  \{A_r(\omega)|m\ |\ r\in S: \emph{arity}(r)\geq m\}.
\end{displaymath}
The set $\{ D_m(\omega)\ |\ m=1,\ldots,\emph{arity}(S)\}$ contains all the information of all adjacency matrices
$A_r(\omega)$, and is therefore an equivalent representation of the world $\omega$. 

Now let $\boldi\in \langle n\rangle^m$. Define
\begin{displaymath}
  D_{\boldi}(\omega) := D_m(\omega \downarrow \boldi).
\end{displaymath}
${\cal T}_m$ is the space of possible values for $ D_{\boldi}(\omega)$ with $\omega\in\Omega$ and
$\boldi\in  \langle \Nset\rangle^m$.
Recall that the projection operation $\omega \downarrow \boldi$ involves a re-naming of the elements of
$\boldi$ as $1,\ldots,m$.
Thus, for example, for the $\omega$ of Figure~\ref{fig:worlddefs} we have that
$D_{(1,3)}(\omega) $ is the element of ${\cal T}_2$ that is graphically represented as
$1 \circ\!\! \rightarrow\!\!   \circ 2$ in Figure~\ref{fig:worlddefs}. 

\section{Additional Examples}

\begin{example}
  Consider the 'bipart'  distribution of Table~\ref{tab:wdistributions}. This is a domain-sampling
  distribution and has an
  AHK$^-$ representation given by functions $f^1, f^2$ as follows.
  Here $\emph{arity}(S)=2$, so AHK models
  only are over variables $U_{\boldi},D_{\boldi}$ with $\boldi\leq 2$. The state space ${\cal T}_1$ contains the
  two states ``self-loop'' and ``no self-loop''. Since self-loops have probability zero in 'bipart', we simply
  let $f^1$ be the constant function that returns ``no self-loop'' regardless of the input $U$-variables.
  The state space ${\cal T}_2$ contains the four states 
  $1\bullet\!\! {\white\rightarrow} \!\! \bullet 2$,
  $1\bullet\!\! \rightarrow\!\! \bullet 2$, $1\bullet\!\! \leftarrow \!\! \bullet 2$,
  and $1\bullet\!\! \leftrightarrow \!\! \bullet 2$, of which only the first and last have non-zero probability.
  To represent this distribution, we let the $U_i$ encode the partition of the domain into two (equal) parts, and
  then define for $i<j$:
  \begin{multline*}
    f^2(U_i,U_j):= \\
    \left\{
      \begin{array}{ll}
        i\bullet\!\! \leftrightarrow \!\! \bullet j & \mbox{if}\ U_i<0.5<U_j\ \mbox{or}\  U_j<0.5<U_i \\
        i\bullet\!\! {\white\leftrightarrow} \!\! \bullet j & \mbox{otherwise}
      \end{array}
      \right.
  \end{multline*}
  Thus, $ f^2$ here neither makes use of $U_{\emptyset}$, nor of $U_{(i,j)}$. If instead of a complete bipartite
  graph we want to model that an (undirected) edge between nodes of different components only exists with a
  certain probability $p<1$, we have to add to the first case in the definition of $f^2$ the condition
  $U_{(i,j)}<p$. It is clear that in a similar manner one can represent arbitrary stochastic block models.
\end{example}

\section{Alternative Sampling Model}
\label{sec:altsampling}

The statistical frequence distribution ${P}^{(k)}(\cdot|\omega)$ defined in Section~\ref{sec:worldletfreqs} is
based on drawing a random ordered sample, and labeling the elements in the sample according to the order in
which they were drawn. For the use in the proof of Theorem~\ref{theo:extendable2} we here introduce
a slightly different sampling model, in which we draw an unordered sample, and  elements are relabeled
according to the order of their original labels. A random $k$-world sampled from a world $\omega\in\Omegan$ then is
given by one of the $\binom{n}{k}$ subesets $\{i_1,i_2,\cdots,i_k \}\subseteq [n]$ where w.l.o.g. we can
assume $i_1<i_2<\cdots<i_k$. The substructure induced by this subset defines a worldlet over $[k]$ via
the identification $i_j\mapsto j$. We denote the resulting sampling distribution by $\hat{P}^{(k)}(\cdot|\omega)$.
Note that for this sampling model we need to assume that we can observe the labels of the randomly drawn
elements from $\omega$ in order to determine their order and to define the re-labeling. Observe, too, that
$\hat{P}^{(k)}(\cdot|\omega)$ in general is not exchangeable: if, e.g. $\omega$ is a linear chain
$1\rightarrow 2 \rightarrow 3 \cdots \rightarrow n$, then
$\hat{P}^{(k)}( 1\circ\!\!\rightarrow\!\! \circ2  | \omega)=(n-1)/\binom{n}{2} \neq
\hat{P}^{(k)}( 1\circ\!\!\leftarrow\!\! \circ2  | \omega)=0$.

Using  $\emph{Iso}(\omega)\subset \Omegan$ to denote the isomorphism class of a world $\omega\in\Omegan$, we obtain
the following relationship between the sampling distributions $P^k$ and $\hat{P}^{(k)}$:
  \begin{equation}
    \label{eq:PandPhat}
    {P}^{(k)}(\omega'|\omega)= \frac{1}{|\emph{Iso}(\omega')|}\sum_{\omega''\in\emph{Iso}(\omega')}  \hat{P}^{(k)}(\omega''|\omega).
  \end{equation}
  If $\hat{P}(\cdot|\omega)$ is exchangeable, then $\hat{P}(\cdot|\omega) ={P}(\cdot|\omega) $.

\section{Proofs for Section~\ref{sec:reptheo}}

In the following proofs we have to consider
  operations on different sets that are defined by permutations of $[n]$. We here list in detail the pertinent
  definitions.

  Let $n\in\Nset$. A permutation  $\pi$ of $[n]$ operates on the  following sets as follows:
  \begin{description}
  \item[A] on $[\Nset]_{\neq}^n$: for $\boldk=(k_1,\ldots,k_n)\in [\Nset]_{\neq}^n$ define
    $\pi \boldk=(k_{\pi(1)},\ldots,k_{\pi(n)})$.
    \item[B]  on $[n]^k$ with $k\geq 1$: for $\boldi=(i_1,\ldots,i_k)\in [n]^k$ define
      $\pi\boldi=(\pi(i_1),\ldots,\pi(i_k))$.
    \item[C] on $\Omegan$: let $\omega\in\Omegan$. $\pi\omega$ then is the world in which for
      a $k$-ary $r\in S$: $A_r(\pi\omega)(\boldi)=1$ iff $A_r(\pi^{-1}\boldi)(\omega)=1$
      ($\boldi\in[n]^k$).
    \item[D] on ${\cal T}^n$: let $t\in {\cal T}^n$ and $r\in S$ with $\emph{arity}(r)=k\geq n$.
      Define $\pi t$ by:
      $(A_r|n)(\boldi)=1$ in $\pi t$ iff  $(A_r|n)(\pi^{-1}\boldi)=1$ in $ t$ ($\boldi\in [n]^k_n$).
      Intuitively, $\pi t$
      is obtained from $t$ by re-labeling the nodes of $t\in {\cal T}^n$ according to $\pi$
      (cf. Figure~\ref{fig:worlddefs}).
    \item[E] on ${2^{[n]}}$:   for $\boldi=\{i_1,\ldots,i_l\}\subseteq [n]$ define
      $\pi\boldi:=\{\pi i_1,\ldots,\pi i_l\}$
    \item[F] on $[2^n]$: for $\boldi\in 2^{[n]}$ let $\emph{Idx}(\boldi)$ be the index of $\boldi$ in the
      lexicographic ordering of $2^{[n]}$. Then, for $1\leq j\leq 2^n$:
      $\pi j:=\emph{Idx}(\pi \emph{Idx}^{-1}(j)  )$, where $\pi \emph{Idx}^{-1}(j) $ is defined according to
      ${\bf E}$.
    \item[G] on $[0,1]^{2^n}$: let $\boldx = (x_1,x_2,\ldots,x_{2^n}) \in [0,1]^{2^n} $. Then
      $\pi\boldx:=(x_{\pi^{-1}1},\ldots,x_{\pi^{-1}2^n})$.
  \end{description}
  According to the above definitions, an expression $\pi\boldk$ is potentially ambiguous, and depends
  on whether $\boldk$ is seen as an element of $[\Nset]_{\neq}^n$, $[n]^k$, or ${2^{[n]}}$. In the following,
  the correct reading of such expressions, if not spelled out explicitly, will be implied by the
  context in which the expression appears.

  Now let $\pi$ be a permutation of $[n]$, let $m<n$, and $\boldi\in\langle n\rangle^m$. Then
  $\boldi=(i_1,\ldots,i_h)$ induces
  a permutation $\pi_{\boldi}$ of $[m]$ as follows:
  \begin{equation}
    \label{eq:pii}
    \pi_{\boldi}: h \mapsto \emph{Idx}(\pi(i_h))\hspace{10mm} (1\leq h\leq m),
  \end{equation}
  where $ \emph{Idx}(\pi(i_h))$ is the index of $\pi(i_h)$ in the ordering of the set $\{\pi(i_h)|1\leq h\leq m   \}$.
  The permutation $\pi_{\boldi}$ then again induces operations {\bf A}-{\bf F} on
  $[\Nset]^m_{\neq}$, \ldots, $[0,1]^{2^m}$ according to the
  previous definitions.

  \begin{example}
    \label{ex:inducedperm}
  Let $n=4$ and $\pi: 1\mapsto 3$, $2\mapsto 1$, $3\mapsto 4$, $4\mapsto 2$. Let $m=2$ and
  $\boldi=(1,4)$. Then $\pi\boldi =(2,3)$, and $\pi_{\boldi}: 1\mapsto 2$, $2\mapsto 1$. Note
  that $\pi\boldi$ being just a set (ordered tuple) does not contain all information on
  how $\pi$ is defined on the components of $\boldi$. However, jointly, $\pi\boldi$ and
  $\pi_{\boldi}$ are sufficient to reconstruct the restriction of $\pi$ to the components of $\boldi$.
\end{example}

  \begin{proofof}{of Proposition~\ref{prop:ahkprojective}}
    Let $S$ be a signature with $\emph{arity}(S)=a$. 
    Let $n\in\Nset$. Let $\omega\in\Omegan$ and $\pi$ a permutation of $[n]$. We have to show
    that
    \begin{equation}
      \label{eq:local050}
      \boldf\downarrow[n](\omega)=\boldf\downarrow[n](\pi\omega).
    \end{equation}
    Let $\boldU_{[n]}$ denote the set of $U_{\boldi}$ variables with $\boldi\in \langle n\rangle^m$
    ($m\leq a$). We can then define
    \begin{equation}
      \label{eq:deffomega}
      \boldf^{-1}(\omega):= \{\boldu\in [0,1]^{\boldU_{[n]}}: \boldf(\boldu)=\omega\}
    \end{equation}
    where $\boldf(\boldu)=\omega$ is shorthand for: for all
    $m\leq a, \boldi=(i_1,\ldots,i_m)\in \langle n\rangle^m$:
    \begin{equation}
      \label{eq:local055}
      f^m(\boldu_{\boldi}) = D_m(\omega\downarrow\boldi),
    \end{equation}
    where $\boldu_{\boldi}=(u_{\emptyset},u_{i_1},\ldots,u_{(i_1,\ldots,u_m)})$. 
    We now need to verify that
    \begin{equation}
      \label{eq:local060}
      \boldf({\boldu})=\omega \Leftrightarrow
      \boldf(\pi{\boldu})=\pi\omega,
    \end{equation}
    where $\pi\boldu$ is defined as the  operation of $\pi$ on $[0,1]^{2^n}$ (case {\bf G}) by
    viewing $\boldu\in[0,1]^{\boldU_{[n]}}$ as an elemenent of $[0,1]^{2^n}$ via the lexicographic
    ordering of the $u_{\boldi}\in\boldu$. 
    This then proves (\ref{eq:local050}) because 
    the joint distribution of the $U_{\boldi}$ variables is invariant under the permutation $\pi$.
    The right-hand side of (\ref{eq:local060}) can be re-written in the form of (\ref{eq:local055}) as:
    for all $m\leq a$, $\boldi\in\langle n\rangle^m$:
    \begin{equation}
      \label{eq:local070}
      f^m((\pi\boldu)_{\boldi})=D_m(\pi\omega\downarrow\boldi)
    \end{equation}
    Since $\pi$ operates as a bijection on $ \langle n\rangle^m$, we can change the variable
    quantified over from  $\boldi$ to $\pi\boldi$, and obtain the equivalent condition that for all
    $\boldi$
      \begin{equation}
      \label{eq:local073}
      f^m((\pi\boldu)_{\pi\boldi})=D_m(\pi\omega\downarrow\pi\boldi)
    \end{equation}
    Now with
    \begin{eqnarray}
        \label{eq:local075}
     & &   (\pi\boldu)_{\pi\boldi} = \pi_{\boldi}\boldu_{\boldi}\\
         \label{eq:local078}
    & &  D_m(\pi\omega\downarrow\pi\boldi)=\pi_{\boldi} D_m(\omega\downarrow\boldi)
    \end{eqnarray}
    and substituting (\ref{eq:local055}) in the right-hand side of (\ref{eq:local078}) we obtain that
    (\ref{eq:local060}) is equivalent that for all $\boldi$ and all $t\in {\cal T}^m$
    \begin{equation}
      \label{eq:local080}
       f^m(\pi_{\boldi}\boldu_{\boldi})=\pi_{\boldi} f^m(\boldu_{\boldi}),
    \end{equation}
    which is exactly the permutation equivariance of $f^m$.

    The projectivity of the family $(\boldf\downarrow [n])_n$ follows directly from the graphical
    structure of the AHK model.
  \end{proofof}

  \begin{example}
    Let $n$ and $\pi$ as in Example~\ref{ex:inducedperm}. Let $S$ contain a single binary relation symbold, and
    let $\omega\in\Omega^{(4)}$ be the world with edges $(1,2),(2,4),(4,1)$. Then $\pi\omega$ is the world with
    edges $(1,2),(2,3),(3,1)$. Let
    \begin{displaymath}
      \boldu=(u_{\emptyset},u_1,\ldots,u_4,u_{(1,2)},\ldots,u_{(2,3,4)},u_{(1,2,3,4)})\in [0,1]^{2^4}
    \end{displaymath}
    such that $\boldf(\boldu)=\omega$. In particular for $\boldi=(1,4)$:
    \begin{displaymath}
      f^2(\boldu_{\boldi})=f^2((u_{\emptyset},u_1,u_4,u_{(1,4)})= 1\circ \leftarrow \circ 2.
    \end{displaymath}
    Then $\pi\boldi$ and $\pi_{\boldi}$ are as in Example~\ref{ex:inducedperm},  and
    \begin{displaymath}
      (\pi\boldu)_{\pi\boldi}=(u_{\emptyset},u_4,u_1,u_{(1,4)})=\pi_{\boldi}(\boldu_{\boldi}).
    \end{displaymath}
    Permutation equivariance of $f^2$ then gives
    \begin{multline*}
      f^2((u_{\emptyset},u_4,u_1,u_{(1,4)})) = \\
      \pi_{\boldi} f^2((u_{\emptyset},u_1,u_4,u_{(1,4)})) = 1\circ \rightarrow \circ 2. 
    \end{multline*}  
  \end{example}

\begin{proofof}{of Theorem~\ref{theo:extendable2}}
  
    $({\bf A})\Rightarrow ({\bf E})$:
    The proof relies on adaptations of concepts and results introduced for the study of graph 
    limits~\cite{borgs2008convergent,diaconis2007graph}, starting with the definition of the function
    $\tauind$ as follows: let ${\cal O}:=[0,1]^{\Omega}$ (to be thought of as the space of probability
    assignments to all possible worlds). Then let
\begin{equation}
  \label{eq:taufunctional}
  \begin{array}{lll}
    \tauind: & \Omega   & \rightarrow {\cal O} \\
    & \omega & \mapsto (\Pn(\omega'|\omega))_{\omega'\in\Omegan,n\geq 1}
   \end{array}
 \end{equation}
 where we define $\Pn(\omega'|\omega)=0$ when the size of  $\omega$ is $<n$. Thus,
 for any $\omega\in\Omega$ the vector $\tauind(\omega)$ has only finitely many non-zero components. 
For $o\in{\cal O}$ we denote with  $o[\omega']$ the $\omega'$ component of  $o$, and with 
$o[\Omegan]$ the tuple of components for $\omega'\in \Omegan$.
Let $(\omega_i)_{i\in \Nset}$ be an enumeration of $\Omega$. The space ${\cal O}$ then can
be equipped with the metric $d(o,o'):=\sum_i \frac{1}{2^i} d(o[\omega_i],o'[\omega_i])$, where
$d$ on the right-hand side is the standard Euclidean metric on $[0,1]$. With this metric,
${\cal O}$ is a compact metric space.

Now assume that $Q^{(k)}$ is a domain sampling distribution. Let $\epsilon_n\rightarrow 0$ ($n\rightarrow\infty$).
For each $n$, let $\omega_n$ be such that (\ref{eq:realizable}) holds with $\omega=\omega_n$ and 
$\epsilon=\epsilon_n$, and such that $\omega_n\in \Omega^{(n')}$ with $n'\geq n$. 
The sequence $(\tauind(\omega_n))_n$ contains a convergent sub-sequence. Let $o^* \in{\cal O}$ be 
its limit. We claim that $o^*$ defines a projective family that extends $Q^{(k)}$:

\begin{itemize}
\item For each $m$: $o^*[\Omegam]$ is a probability distribution on $\Omegam$.
  Let ${\cal O}^{(m)}_{\emph{prob}}$ be
  the subset of $o\in {\cal O}$ for which $o[\Omegam]$ is a probability distribution. Then
  ${\cal O}^{(m)}_{\emph{prob}}$ is 
  closed,  and $\tauind(\omega_n)\in {\cal O}^{(m)}_{\emph{prob}}$ for all $n>m$.
Thus, also $o^*\in {\cal O}^{(m)}_{\emph{prob}}$.
\item For $m'>m$: $o^*[\Omega^{m'}]\downarrow m = o^*[\Omegam]$. Let
${\cal O}^{(m,m')}_{\emph{proj}}$ be
the subset of $o\in {\cal O}$ for which  $o[\Omega^{m'}]\downarrow m = o[\Omegam]$. Again,
this is a closed set,  $\tauind(\omega_n)\in {\cal O}^{(m,m')}_{\emph{proj}}$ for all $n\geq m'$,
and therefore also  $o^*\in {\cal O}^{(m,m')}_{\emph{proj}}$.
\item For each $m\geq k$: $o^*[\Omegam]\downarrow k = Q^{(k)}$: this follows from the 
definition of the sequence $\omega_n$.
\end{itemize}
({\bf B})$\Rightarrow$ ({\bf A}):
Let $\boldf$ be an AHK$^-$ model that represents $\Qk$. 
Let $n\geq k$ and consider $\boldf\downarrow [n]$.
Let $\boldD$ be the vector of all $D_{\boldi}$-variables with  $\boldi\subset [n]$, i.e.,
$\boldD$ represents a random $\omega\in\Omegan$ drawn according to $\boldf\downarrow [n]$.
For any $\boldk\in\langle n\rangle^k$ let
$\boldD_{\boldk}=(D_{\boldi})_{\boldi\in\langle n\rangle^m:\boldi\subseteq\boldk;m=1,\ldots,a}$ be the
vector of variables that represent the worldlet induced by $\boldk$. Now consider a fixed $\omega'\in\Omegak$ and the random variables
$\indicator{\boldD_{\boldk}=\omega'}$ where $\indicator{}$ is the indicator function.
Since $\boldf$ represents $\Qk$, we have that
\begin{equation}
  \label{eq:local90}
  E[\indicator{\boldD_{\boldk}=\omega'}]=\Qk(\omega').
\end{equation}
Taking the average over all $\boldk\in\langle n\rangle^k$, we obtain the random variable
\begin{equation}
  \label{eq:local100}
  \frac{1}{\binom{n}{k}}\sum_{\boldk\in\langle n\rangle^k} \indicator{\boldD_{\boldk}=\omega'}=
    \hat{P}^{(k)}(\omega'|\boldD),
  \end{equation}
  with $\hat{P}$ as introduced in Appendix~\ref{sec:altsampling}. 
   From (\ref{eq:local90}) it
  follows that still
  \begin{equation}
    \label{eq:local105}
    E[\hat{P}^{(k)}(\omega'|\boldD)]=\Qk(\omega').
  \end{equation}
  
The variables
$\boldD_{\boldk}$ and $\boldD_{\boldk'}$  are independent according to $\boldf\downarrow [n]$
whenever $\boldk\cap\boldk'=\emptyset$\
(note that this is not the case in the AHK model, where the $U_{\emptyset}$ variable can induce a
dependency).
The family of random variables $\indicator{\boldD_{\boldk}=\omega'}$ ($\boldk\in \langle n\rangle^k$)
has the same (weak) dependence structure as found in U statistics~\cite{hoeffding1963probability}.
The probability bound given by Hoeffding~\shortcite[Equation 5.7]{hoeffding1963probability} and extended to more
general scenarios by Janson~\shortcite{janson2004large}
(cf. Corollary 2.2 and Example 4.1 in~\cite{janson2004large}) provide the following bound for
the distribution of the random variable defined by (\ref{eq:local100}): for $t>0$
\begin{equation}
  \label{eq:local110}
  P_{\boldD}(  \hat{P}^{(k)}(\omega'|\boldD) \geq \Qk(\omega')+t  ) \leq e^{-2 \lfloor \frac{n}{k} \rfloor t^2},
\end{equation}
where $P_{\boldD}$ denotes the distribution defined by $\boldf\downarrow[n]$ on $\boldD$.
The same bound holds for $P_{\boldD}(  \hat{P}^{(k)}(\omega'|\boldD) \leq \Qk(\omega')-t  )$.

Now let $\epsilon>0$ as in Definition~\ref{def:dsrealizable} be given. Then, for any $n\in\Nset$
\begin{equation}
  \label{eq:local120}
  2\cdot|\Omegak|\cdot e^{-2 \lfloor \frac{n}{k} \rfloor \epsilon^2}
\end{equation}
is an upper bound for the probability that for some $\omega'\in\Omegak$ the frequency
$\hat{P}^{(k)}(\omega'|\boldD)$ is outside the bounds $\Qk(\omega')\pm \epsilon$. For all sufficiently
large $n$, this bound is $<1$, which, in particular means that for all sufficiently large $n$ there
exist $\omega_n\in\Omegan$ with
\begin{equation}
  \label{eq:local130}
  \hat{P}^{(k)}(\omega'|\omega_n)=\Qk(\omega')\pm \epsilon
\end{equation}
for all $\omega'\in\Omegak$. 
Since $\Qk$ is exchangeable, it follows from (\ref{eq:PandPhat}) that (\ref{eq:local130}) also
holds when $\hat{P}^{(k)}$ is replaced by $ {P}^{(k)}$.


    ({\bf C})$\Rightarrow$ ({\bf E}): the set of  projective extendable $\Qk$ is convex.
    The implication then follows from $({\bf A})\Rightarrow ({\bf E})$.
    
    

 
  
  ({\bf D})$\Rightarrow$ ({\bf C}): The set of  extendable distributions is the closed convex
  set ${\cal P}^{(k)}:=\cap_n \Delta^{(k)}_{n}$. Since ${\cal P}^{(k)}$ has finite dimension, each point
  of ${\cal P}^{(k)}$ is a finite convex combination of its extreme points.
  Let $p$ be an extreme point of  ${\cal P}^{(k)}$.
  We show that $p$ is
  a domain-sampling distribution.
  Each $\Delta^{(k)}_{n}$ is a polytope whose vertices are statistical frequency
  distributions $\Pk(\cdot|\omega)$ for some $\omega\in\Omegan$.
  With Lemma~\ref{lem:convex} it follows that $p$ is
  the limit of a convergent sequence of distributions $\Pk(\cdot|\omega_n)$, and thus
  a domain sampling distribution.
  
  
  ({\bf E})$\Rightarrow$ ({\bf F}):
  Let $\Omega^{(\Nset)}$ denote the set of infinite worlds over the node set $\Nset$.
  Let $\{\Qn| n\geq 1\}$ be a projective family extending $\Qk$.
  By the standard Kolmogorov existence theorem and the exchangeability of the $\Qn$,
  there exists a distribution $Q^{(\Nset)}$ on
  $\Omega^{(\Nset)}$ such that $Q^{(\Nset)}\downarrow \boldk =\Qn$ for any $\boldk\in [\Nset]_{\neq}^n$
  ($n\geq 1$).
  Let $W_{\Nset}$ be a $Q^{(\Nset)}$-distributed random variable,
  and for  $\boldk\in [\Nset]_{\neq}^m$ with $m\in [a]$  
  let
  \begin{equation}
    \label{eq:DkDef}
    \tilde{D}_{\boldk}:=D_m(W_{\Nset}\downarrow\boldk).
  \end{equation}
  The variables $( \tilde{D}_{\boldk})_{\boldk}$ then form
  an $a$-dimensional exchangeable array in the sense of~\cite{kallenberg2006probabilistic}.
  Even though we will later identify the variables $\tilde{D}_{\boldk}$ with the 
  $D_{\boldk}$-variables of our target AHK representation, we distinguish them for now via the $\tilde{D}$ notation.
  According to the representation
  theorm~\cite[Lemma 7.25]{kallenberg2006probabilistic} the joint distribution of the $(\tilde{D}_{\boldk})_{\boldk}$ has a
  representation given by
  \begin{itemize}
  \item a family of i.i.d. random variables $\{U_{\boldi}|  \boldi\in \langle \Nset \rangle^m, m=0,\ldots,a\}$ as
    described in Definition~\ref{def:ahkmodel};
  \item functions
    \begin{displaymath}
      {f}^m:[0,1]^{2^m}\rightarrow\Omega^{(m)}\ (m=1,\ldots,a)
    \end{displaymath}
  \end{itemize}
  such that for $\boldk$ of length $m$:
  \begin{equation}
    \label{eq:Dkvars}
    \tilde{D}_{\boldk}={f}^m(\boldU_{\boldk}) 
  \end{equation}
  where the arguments $\boldU_{\boldk}$ consist of all $U_{\boldk'}$ with $\boldk'\subseteq \boldk$ enumerated
  according to the lexicographic ordering of the indices of $\boldk'$ in $\boldk$.
  In detail: if $\boldk=(k_1,\ldots,k_m)$, $\boldk'=\{k_{i'_1},\ldots,k_{i'_{d'}}\}$,
  $\boldk''=\{k_{i''_1},\ldots,k_{i''_{d''}}\}$ with $i'_1<\cdots<i'_{d'}$ and
  $i''_1<\cdots<i''_{d''}$, then $U_{\boldk'}$  precedes $U_{\boldk''}$ in $\boldU_{\boldk}$ if
  $(i'_1,\ldots,i'_{d'})$ precedes $(i''_1,\ldots,i''_{d''})$ lexicographically.
  Note that
  when $\boldk$ is ordered as in Definition~\ref{def:ahkmodel}, the lexicographic ordering
  of subsets $\boldk',\boldk''\subseteq\boldk$ according to their indices in $\boldk$ corresponds to the
  lexicographic ordering of the tuples $\boldk',\boldk''$ themselves.

  To transform this representation into the form stated in Definition~\ref{def:ahkmodel} we
  exploit the fact that the family $(\tilde{D}_{\boldk})_{\boldk}$ not only is exchangeable, but also exhibits
  certain deterministic permutation invariance relationships.
  
  Now let $\pi$ be a permutation of $[m]$. Let $\boldk\in[\Nset]_{\neq}^m$. For the
  variables $\tilde{D}_{\boldk}$ of (\ref{eq:Dkvars}) we then have
  \begin{equation}
    \label{eq:DpikDk}
     \tilde{D}_{\pi\boldk}= D_m(W_{\Nset}\downarrow \pi\boldk) =  \pi(\tilde{D}_{\boldk}).
   \end{equation}
   Thus, it is sufficient to consider the $\tilde{D}_{\boldk}$ for ordered $\boldk$, and we now let
   for $\boldk\in\langle\Nset\rangle^m\subset [\Nset]^m_{\neq}$:
   \begin{equation}
     \label{eq:defproofDk}
     D_{\boldk}:=\tilde{D}_{\boldk}.
   \end{equation}
  According to (\ref{eq:Dkvars})
  \begin{displaymath}
    \begin{array}{l}
      \tilde{D}_{\pi\boldk}= f^m( \boldU_{\pi\boldk}  ) \\
      \pi(\tilde{D}_{\boldk}) = \pi  f^m( \boldU_{\boldk}  ) 
    \end{array}
  \end{displaymath}
  Together with
   \begin{displaymath}
    \boldU_{\pi\boldk}=\pi\boldU_{\boldk},
  \end{displaymath}
  then (\ref{eq:DpikDk}) gives the permutation equivariance of $f^m$:
  \begin{displaymath}
    f^m(\pi\boldU_{\boldk})=\pi(f^m(\boldU_{\boldk})).
  \end{displaymath}

  
  ({\bf A})$\Rightarrow$ ({\bf B}): Let $\Qk$ be a domain sampling distribution.
  From ({\bf A})$\Rightarrow$ ({\bf E} )$\Rightarrow$ ({\bf F}) we know
  that $\Qk$ has an AHK representation $\boldf$. For the purpose of this proof we have to
  be careful to properly distinguish between the functions $\boldf$, and the distribution
  they define. We write $P_{\boldf}$ for the distribution defined by $\boldf$ (undoing the
  slight abuse of notation introduced after the statement of Definition~\ref{def:ahkmodel}).
  When, in the following, we refer to the marginal distribution of the continuous
  variable $U_{\emptyset}$, then expressions of the form $P_{\boldf}(U_{\emptyset}=x|A)$ stand for the
  value at $x\in[0,1]$ of the density function representing the conditional distribution
  of $U_{\emptyset}=x$ given the event $A$. 
  Since $\Qk$ is a domain sampling distribution we have that for tuples 
  $\boldk,\boldk'\in\langle \Nset \rangle^m$ with $\boldk\cap\boldk'=\emptyset$ the
  variables $D_{\boldk},D_{\boldk'}$ are independent:
  \begin{equation}
    \label{eq:local400}
    P_{\boldf}(D_{\boldk'}|D_{\boldk})=P_{\boldf}(D_{\boldk'}).
  \end{equation}
  We first show that the $D_{\boldk}$ are independent of $U_{\emptyset}$. Assume otherwise. Then there exists
  a $t\in{\cal T}^m$ such that $P_{\boldf}(D_{\boldk}=t|U_{\emptyset}=x)$ is not almost surely constant
  for $x\in [0,1]$. 
  We then obtain a contradiction to (\ref{eq:local400}) by first expanding:
  \begin{multline*}
    P_{\boldf}(D_{\boldk'}=t|D_{\boldk}=t)=\\
    \int_{[0,1]} P_{\boldf}(U_{\emptyset}=x|D_{\boldk}=t) P_{\boldf}(D_{\boldk'}=t|U_{\emptyset}=x,D_{\boldk}=t)dx.
  \end{multline*}
  Using the conditional independenc of $D_{\boldk},D_{\boldk'}$ given $U_{\emptyset}$, this simplifies to:
  \begin{multline*}
      \int_{[0,1]} P_{\boldf}(U_{\emptyset}=x|D_{\boldk}=t) P_{\boldf}(D_{\boldk'}=t|U_{\emptyset}=x)dx.
    \end{multline*}
    With Bayes's rule and $P_{\boldf}(U_{\emptyset}=x)=1$ this becomes:
   \begin{multline*}
    \int_{[0,1]} \frac{P_{\boldf}(D_{\boldk}=t|U_{\emptyset}=x)}{P_{\boldf}(D_{\boldk}=t)} P_{\boldf}(D_{\boldk'}=t|U_{\emptyset}=x)dx, \\
  \end{multline*}
  and using $P_{\boldf}(D_{\boldk}=t|U_{\emptyset})=P_{\boldf}(D_{\boldk'}=t|U_{\emptyset})$ and
  $P_{\boldf}(D_{\boldk}=t)=P_{\boldf}(D_{\boldk'}=t)$ we finally obtain:
  \begin{equation}
    \label{eq:local430}
    P_{\boldf}(D_{\boldk'}=t|D_{\boldk}=t)=\int_{[0,1]} \frac{P_{\boldf}(D_{\boldk'}=t|U_{\emptyset}=x)^2}{P_{\boldf}(D_{\boldk'}=t)}dx 
  \end{equation}
  Expanding also the right-hand side of (\ref{eq:local400}) we can write:
  \begin{equation}
    \label{eq:local435}
     P_{\boldf}(D_{\boldk'}=t)=\int_{[0,1]} P_{\boldf}(D_{\boldk'}=t|U_{\emptyset}=x)dx.
   \end{equation}
   We then obtain
   \begin{equation}
     \label{eq:local440}
      P_{\boldf}(D_{\boldk'}=t|D_{\boldk}=t)>P_{\boldf}(D_{\boldk'}=t)
   \end{equation}
   by multiplying both (\ref{eq:local430}) and (\ref{eq:local435}) with $ P_{\boldf}(D_{\boldk'}=t)$, and
   an application of Jensen's inequality to the random variable
   $P_{\boldf}(D_{\boldk'}=t|U_{\emptyset})$ and the convex function $x\mapsto x^2$. The inequality is strict, because
   of the strict convexity of the square function, and the assumption that $P_{\boldf}(D_{\boldk'}=t|U_{\emptyset})$
   is not almost surely constant.

   The above argument can be carried out in exactly the same manner (only with a heavier load of notation) to
   also show that every finite family $D_{\boldk_1},\ldots,D_{\boldk_n}$
   ($\boldk_i\in \langle \Nset\rangle^{m_i}, m_i\leq a, n\geq 1$) is independent of $U_{\emptyset}$. It then follows that in the AHK model representing $\Qk$ the
   node $U_{\emptyset}$ is redundant and can be
   eliminated. It remains to show that the conditional distributions of the $D_{\boldi}$ on the remaining
   $U_{\boldi'}$ variables $(\emptyset\neq\boldi'\subseteq\boldi)$ can still be represented by  deterministic
   measurable, permutation-equivariant functions $f^m$.

   Let $A(D_{\boldk_1},\ldots,D_{\boldk_n})\subseteq [0,1]$ be such that $P(A(D_{\boldk_1},\ldots,D_{\boldk_n}))=1$ (under
   the uniform distribution on $[0,1]$), and
   \begin{equation}
     \label{eq:local450}
     P_{\boldf}(D_{\boldk_1},\ldots,D_{\boldk_n})= P_{\boldf}(D_{\boldk_1},\ldots,D_{\boldk_n}|U_{\emptyset}=x)
   \end{equation}
   for all $x\in A(D_{\boldk_1},\ldots,D_{\boldk_n}) $. Since there are only countably many finite families of
   $D_{\boldk_i}$, also the intersection of the sets $A(D_{\boldk_1},\ldots,D_{\boldk_n})$ for all such families has
   probability 1, and, in particular, there exists an $x^*\in[0,1]$ such that (\ref{eq:local450}) holds with
   $x=x^*$ for all families. For every $m\leq a$ now define
   \begin{displaymath}
     \tilde{f}^m(x_1,\ldots,x_{2^m-1}):= {f}^m(x^*,x_1,\ldots,x_{2^m-1}).
   \end{displaymath}
   The $\tilde{f}^m$  (in conjunction with uniform distributions on all $U_{\boldi}$ ($\boldi\neq\emptyset$))
   then represent the original distribution $P_{\boldf}$ (marginalized on all variables other than $ U_{\emptyset}$),
   and inherit the permutation equivariance of the $f^m$. Note that the operation of permutations $\pi$ on
   $[0,1]^{2^m}$ leaves the first component  invariant, and therefore can also be directly seen as
   permutations on the domains  $[0,1]^{2^m-1}$ of the $\tilde{f}^m$.
   
  ({\bf F})$\Rightarrow$ ({\bf E}): this follows from Proposition~\ref{prop:ahkprojective}. 
\end{proofof}

\begin{lemma}
  \label{lem:convex}
  For $n\geq 1$ let ${\cal P}_n$ be a polytope in $\Rset^k$, such that ${\cal P}_{n+1}\subseteq {\cal P}_n$ for
  all $n$. Let ${\cal P}:=\cap_n {\cal P}_n$, and $p$ an extreme point of ${\cal P}$. Then there exists for each
  $n$ a vertex $p_n$ of  ${\cal P}_n$ such that $\lim_{n\rightarrow\infty} p_n =p$.
\end{lemma}

\begin{proof}
  It follows from Straszewicz's theorem that it is sufficient to consider the case that $p$ is an
  exposed point of ${\cal P}$, i.e., there exists a hyperplane $H$, such that ${\cal P}\cap H =\{p\}$.
  Let $H^+$ be the closed half-space defined by $H$ for which also ${\cal P}\cap H^+ =\{p\}$.
  For each $n$ then ${\cal Q}_n:={\cal P}_n \cap H^+$ is a polytope, and $\{p\}=\cap_n  {\cal Q}_n$. Thus,
  for any sequence of vertices $q_n\in {\cal Q}_n$ we have $\lim_n q_n = p$. To conclude the proof we
  have to show that each ${\cal Q}_n$ has at least one vertex that is also a vertex of the original ${\cal P}_n$.
  For this we consider two cases: first, assume that ${\cal P}_n\cap H^+\setminus H \neq \emptyset$.
  Then every vertex of ${\cal P}_n$ that lies within $H^+\setminus H $ also is a vertex of
  ${\cal Q}_n$, and there exists at least one such vertex. Second, assume that
  ${\cal P}_n\cap H^+\subset H$. Then $H$ is a supporting hyperplane of ${\cal P}_n$, and all vertices
  of ${\cal P}_n\cap H$ are also vertices of ${\cal P}_n$.
\end{proof}

\section{Proposition~\ref{prop:modularity}: Correction and Proof} 

Proposition~\ref{prop:modularity} as stated in the paper contained a typo and an incorrect
statement at the end. A corrected version of the theorem is as follows:

\begin{proposition}
  \label{prop:modularity_update}
  Let $\boldf$ be an AHK model, $\omega\in\Omegan$ with
  $p:=\boldf\downarrow[n](\omega)>0$.
  Let $O\subset\Omega^{(n+1)}$ be the set of $n+1$ worlds $\omega'$ for which
  $\omega'\downarrow [n]=\omega'\downarrow (1,\ldots,n-1,n+1 ) =\omega$.
  Then $\boldf\downarrow[n](O)\geq p^2$.
\end{proposition}

The corrected typo is a replacement of $\downarrow \{\ldots\}$ by $\downarrow (\ldots)$.
The conditional independence claimed in the original version of the proposition does, in fact,
not even hold in AHK$^-$ models. Proposition~\ref{prop:modularity_update} sharpens the main
statement of the original proposition slightly by providing the quantitative bound $\geq p^2$.
These corrections do not affect  Figure~\ref{fig:modular}, or the application of the proposition to
the '+' distribution of Table~\ref{tab:wdistributions}.

\begin{proof}

  Let $I=\cup_{m=0}^{\emph{arity}(S)} \langle n+1 \rangle^m$, so that 
  the distribution over worlds $\omega'\in\Omega^{(n+1)}$ is modeled by the random variables
  $\{U_{\boldi},D_{\boldi}| \boldi\in I\}$.
  We partition $I$ into 4 subsets as follows:  $I_1$ contains
  the $\boldi\subseteq [n-1]$ (including $U_{\emptyset}$).
  $I_2 $ contains
  the  $\boldi$ which contain $n$, but not $n+1$;  $I_3$ contains the
   $\boldi$ which contain $n+1$, but not $n$, and $I_4$ contain the
   $\boldi$ which contain both $n$ and $n+1$.
   For $k=1,\ldots, 4$ we write $\boldU_k,\boldD_k$ for the sets of variables $U_{\boldi},D_{\boldi}$ with
   $\boldi\in I_k$, and $\boldu_k$ to denote a set of values for $\boldU_k$.
   We can then decompose $\omega'\in\Omega^{(n+1)}$ as 
   \begin{equation}
     \label{eq:omegadecomp}
     \omega'=(\omega^-,\rho_2,\rho_3,\rho_4)
   \end{equation}
   where $\omega^-=\omega'\downarrow [n-1]$ is determined by the variables
   $\boldD_1$, and $\rho_k$ is the sub-structure defined by the variables $\boldD_k$
   ($k=2,3,4$). Unlike the first component $\omega^-$, these sub-structures are not
   possible worlds in their own right.
   Now, let $\omega$ be as given in the proposition. Limiting the decomposition
   (\ref{eq:omegadecomp}) to $\Omegan$, we can write $\omega=(\omega^-,\rho_2)$. From now on
   let $\omega^-,\rho_2$ be the fixed structures defined by $\omega$, and let $\rho_3$ be the
   sub-structure that is isomorphic to $\rho_2$ via the re-naming $n+1\mapsto n$. Then
   $\omega'\in O$ iff the first three components in its decomposition are $(\omega^-,\rho_2,\rho_3)$. 
   
   The variables $\boldD_2$ that define $\rho_2$, for example, are a function
   of $\boldU_1$ and $\boldU_2$. We write $\boldf(\boldu_1,\boldu_2)=\rho_2$ when the values
   $\boldu_1,\boldu_2$ induce the substructure $\rho_2$. Similarly for other substructures.
   Using  again $\indicator{}$ to denote the indicator function, we can now write
   \begin{multline}
     \label{eq:local600}
     p=\int_{[0,1]^{\boldU_1}}\int_{[0,1]^{\boldU_2}}\indicator{\boldf(\boldu_1)=\omega^-}\cdot \\
     \indicator{\boldf(\boldu_1,\boldu_2)=\rho_2}d\boldu_2 d\boldu_1.
     \end{multline}
 and
  \begin{multline}
    \label{eq:local610}
    P(\boldf^{-1}(O)) =
    \int_{[0,1]^{\boldU_1}}\int_{[0,1]^{\boldU_2}}\int_{[0,1]^{\boldU_3}}\indicator{\boldf(\boldu_1)=\omega^-}\cdot \\
    \indicator{\boldf(\boldu_1,\boldu_2)=\rho_2} \cdot
    \indicator{\boldf(\boldu_1,\boldu_3)=\rho_3}d\boldu_3 d\boldu_2\boldu_1.
  \end{multline}
  Since the sub-structures $\rho_2,\rho_3$ are ismorphic and  given by the
  same functional dependency on $\boldu_1,\boldu_2$, respectively $\boldu_1,\boldu_3$, we can define:
  \begin{multline}
    \label{eq:local620}
    F(\boldu_1):= 
    \int_{[0,1]^{\boldU_2}}
    \indicator{\boldf(\boldu_1,\boldu_2)=\rho_2}d\boldu_2 = \\
    \int_{[0,1]^{\boldU_3}}
    \indicator{\boldf(\boldu_1,\boldu_3)=\rho_3}d\boldu_3.
  \end{multline}
  Then (\ref{eq:local600}) becomes
  \begin{equation}
    \label{eq:local630}
    \int_{[0,1]^{\boldU_1}} \indicator{\boldf(\boldu_1)=\omega^-} F(\boldu_1)d\boldu_1,
  \end{equation}
  and (\ref{eq:local610}) becomes
\begin{equation}
    \label{eq:local640}
    \int_{[0,1]^{\boldU_1}} \indicator{\boldf(\boldu_1)=\omega^-} F^2(\boldu_1)d\boldu_1.
  \end{equation}
  Applying Jensen's inequality to the random variable $\indicator{\boldf(\boldu_1)=\omega^-} F(\boldu_1)$ then
  proves the proposition (observing that the square of the indicator function is the indicator function
  itself).
\end{proof}

\bibliographystyle{named}
\bibliography{mplan}

\end{document}